\documentclass[10pt, twocolumn, letterpaper]{article}
\usepackage[margin=0.75in, columnsep=0.25in]{geometry}
\usepackage{amsmath}
\usepackage{amssymb}
\usepackage{hyperref}
\usepackage{natbib}
\usepackage{booktabs}
\usepackage{setspace}
\usepackage{times}
\usepackage{parskip}
\usepackage{titlesec}
\newcommand\samethanks[1][\value{footnote}]{\footnotemark[#1]}
\usepackage{graphicx}
\usepackage{enumitem}
\usepackage{cleveref}
\usepackage{subcaption}
\usepackage{flushend}
\usepackage{dblfloatfix} 

\usepackage{placeins}
\usepackage{authblk}
\titleformat{\section}{\normalsize\bfseries}{\thesection.}{1em}{}
\titleformat{\subsection}{\normalsize\bfseries}{\thesubsection.}{1em}{}
\titleformat{\subsubsection}{\small\itshape}{\thesubsubsection.}{1em}{}
\hypersetup{
    colorlinks=true,
    linkcolor=black,
    citecolor=black,
    urlcolor=black
}
\title{\textbf{The Ethical Decision Head: Operationalizing Normative Ethics
in Autonomous Vehicles via Reinforcement Learning from Human Feedback}}

\author[1]{Thomas Mbrice\thanks{Equal contribution.}}
\author[1]{Ammar Ali\samethanks}
\author[1]{Sami Mian}
\author[1]{Khai Hern Low}
\author[1]{Eric Chen}
\author[1]{Arshia Aghajani}
\author[2]{Wolf Sch\"afer}
\author[2]{Amin Shirangi\thanks{Corresponding author.}}
\affil[1]{Department of Computer Science, Stony Brook University}
\affil[2]{Department of Technology, AI and Society, Stony Brook University}

\date{}
\begin{document}
\maketitle

\begin{abstract}
As autonomous vehicles (AVs) approach Level 4 and Level 5 operational
capability \citep{sae2018}, their onboard decision systems must handle not
only safety-critical locomotion but also their subsequent moral weight. This
paper details the Ethical Decision Head (EDH), a deep reinforcement
learning (RL) framework that encodes ethical reasoning as a differentiable
reward signal, enabling a policy gradient agent to learn morally-aligned
driving behavior in scenarios whose state representation is aligned with the
CARLA simulation environment \citep{dosovitskiy2017}. Two normative
frameworks are instantiated and evaluated: a Utilitarian framework minimizing
total casualties and a Kantian framework enforcing course maintenance as a
categorical imperative. The EDH is trained via Proximal Policy Optimization
(PPO) \citep{schulman2017ppo} against a Bradley-Terry reward model
\citep{bradleyterry1952} learned from pairwise human preference annotations
over 200 collision-imminent scenarios. Results reveal an asymmetry in the
learnability of normative ethical frameworks under human supervision. The
Kantian condition, which reduces to a constant prediction task under the
codebook, serves as a pipeline control: it confirms training stability and
rules out infrastructure failure as an explanation for the utilitarian result.
The Utilitarian agent learned something more unsettling: human raters rewarded
self-sacrifice over casualty minimization, and the model learned that
preference faithfully. This divergence between what humans prescribe in theory
and what they reward in practice suggests that RLHF does not learn ethics as
philosophers define it, but as humans live it.
\end{abstract}

\FloatBarrier
\section{Introduction}

The trolley problem, first posed by Philippa Foot in 1967 \citep{foot1967},
was never intended to be an engineering specification. Yet as autonomous
vehicles approach SAE Level 4 and Level 5 operational capability
\citep{sae2018}, the thought experiment has quietly become one. Every
collision-imminent junction a self-driving system encounters is, in some form,
a trolley problem: a forced choice between outcomes that mathematics can rank
but moral philosophy has never conclusively resolved. \citet{bonnefon2016}
demonstrated in a landmark study that humans hold deeply inconsistent ethical
preferences in exactly these scenarios, favoring utilitarian outcomes in the
abstract while resisting them when applied to themselves or their families.
This inconsistency is not a failure of moral reasoning but rather a feature.

The emergence of Reinforcement Learning from Human Feedback (RLHF) as a
dominant post-training paradigm \citep{christiano2017, stiennon2020} presents
a novel opportunity: rather than encoding ethics as a static reward function,
a brittle and philosophically contentious approach, an agent can learn moral
behavior directly from human preference signals. This raises a question that
is both technical and philosophical: can a machine learning system learn human
ethics the way a human does, not as philosophers prescribe it, but as people
actually practice it?

This paper attempts to answer that question empirically. We introduce the
Ethical Decision Head (EDH), a post-training architectural component that
frames AV ethical decision-making as an RLHF problem, instantiated under two
normative frameworks: Utilitarianism, which prescribes casualty minimization
\citep{bentham1789, mill1863}, and Kantianism, which prohibits the
instrumentalization of rational agents regardless of outcome \citep{kant1785}.
By evaluating both frameworks under identical training conditions, we expose a
fundamental asymmetry in the learnability of classical ethical theory under
human supervision and, in doing so, offer an empirical contribution to a
debate that has until now remained almost entirely theoretical. We implement
this through a PPO-based policy trained against a Bradley-Terry reward model,
evaluated across 200 human-annotated collision scenarios under two training
configurations.

\FloatBarrier
\section{Methodology}

\subsection{Overview and Motivation}

This research operationalizes the problem through the framework of Reinforcement
Learning from Human Feedback (RLHF), built into an architectural component
called the Ethical Decision Head (EDH). RLHF is chosen over static reward
engineering for the property that ethical preferences resist formal
specification \citep{gabriel2020, russell2019}. A hand-crafted reward function
encoding ``minimize casualties'' is straightforward to write but fails to
capture the granularity of human moral intuition: the contextual weightings,
implicit duties, and affective responses that humans bring to morally charged
scenarios. RLHF instead learns this signal directly from human judgment
\citep{christiano2017}, making it a natural fit for a domain where the ground
truth is contested by design.

The EDH directs an underlying AV model in safety-critical scenarios. At each
collision-imminent junction, the agent overrides the underlying AV behavior to
choose among three actions: maintain course, swerve left, or swerve right.
This intervention occurs exclusively when standard braking maneuvers are
insufficient to prevent collision \citep{paden2016}. The EDH operates under one of two ethical
frameworks: a Utilitarian framework that minimizes total casualties, or a
Kantian framework that treats maintaining course as a universal duty, never
using persons as a means to an end \citep{kant1785}. The formal, quantitative
definition of the EDH, encompassing its policy network, reward model, and
training objective, is elaborated in the sections that follow. The qualitative
framing of scenarios, as presented to human annotators for preference
elicitation, is defined alongside the dataset. This framing is intentional;
rather than encoding ethics as a static reward function, this paper argues
that ethical behavior must be learned from human preference signals and
grounded in formal moral structure simultaneously.

\subsubsection{Composite Reward Function}
\label{ssec:composite}

The total reward signal driving policy optimization is a weighted composition
of three terms:

\begin{equation}
  \label{eq:reward_total}
  R_{\text{total}}(s, a) \;=\;
    \alpha \, R_{\text{base}}(s, a)
    + \beta \, R_{\text{learned}}(s, a)
    + \lambda \, R_{\text{explicit}}(s, a)
\end{equation}

The default configuration uses $\alpha = 0.3$ and $\beta = 0.7$, making
learned human preferences the dominant signal. $R_{\text{base}}$ encodes hard
safety constraints, providing a stable dense signal during early training
before the reward model has accumulated sufficient preference data.
$R_{\text{learned}} = R_\psi(s, a)$ is the output of the trained reward
model, capturing the implicit moral intuition of human annotators aggregated
over the preference dataset. The explicit philosophical term
$R_{\text{explicit}}$ is gated by $\lambda \in \{0, 1\}$ and is inactive by
default; it can be enabled during ablation to study the effect of hard-coded
ethical constraints in isolation.

\subsubsection{Explicit Ethical Reward Terms}
\label{ssec:explicit}

When $\lambda = 1$, the following formal encodings of the two normative
frameworks are active.

\paragraph{Utilitarian Penalty.}
The utilitarian component penalizes actions proportionally to expected
casualties:
\[
  R_u(s, a) \;\propto\; -N_{\text{injured}}(s, a)
\]
where $N_{\text{injured}}(s, a)$ is the pedestrian collision count encoded in
the state vector for action $a$ in state $s$, implementing the classical
imperative to minimize aggregate harm \citep{bentham1789, mill1863}.

\paragraph{Kantian Penalty.}
The Kantian component penalizes actions in which the agent instrumentalizes a
pedestrian, that is, deliberately swerves through a pedestrian to reduce the
utilitarian count elsewhere:

\[
  R_k(s, a) \;=\;
  \begin{cases}
    -100 & \text{if } a \in \{\text{swerve-left, swerve-right}\} \\
         & \quad\text{and } N_{\text{injured}}(s,a) > 0 \\
    0    & \text{otherwise}
  \end{cases}
\]

The magnitude $-100$ is chosen to be large relative to any achievable
positive reward, ensuring that deliberate use of pedestrians as obstacles is
never selected by the policy under any weighting configuration \citep{kant1785}.

\subsection{Dataset}

The dataset consists of 200 scenarios constructed according to a codebook,
which acts as a strict rule guide for scenario creation and is included in
this paper as a supplement. While 200 scenarios
represent a constrained sample, the diversity of the codebook-guided
scenario generation process ensures coverage across the key variable
dimensions of the state space. Dataset scale is acknowledged as a limitation
in Section~\ref{sec:limitations}.

\begin{figure*}[!t]
  \centering
  \begin{subfigure}[t]{0.44\textwidth}
    \includegraphics[width=\textwidth]{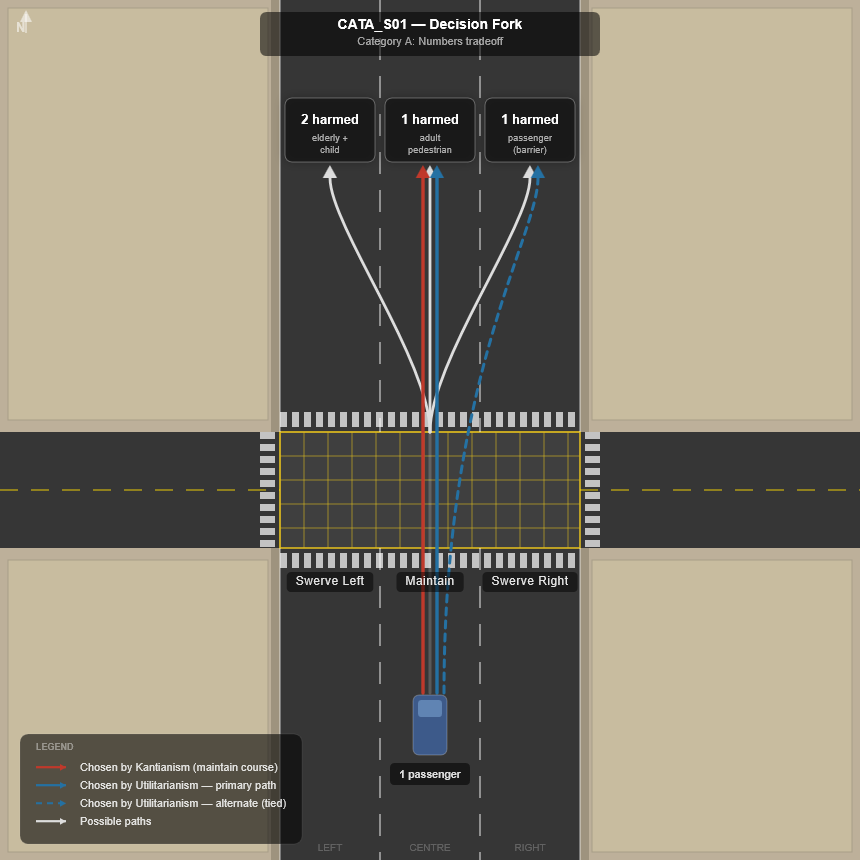}
    \caption{CARLA scenario render for illustrative purposes.}
  \end{subfigure}
  \hfill
  \begin{subfigure}[t]{0.44\textwidth}
    \includegraphics[width=\textwidth]{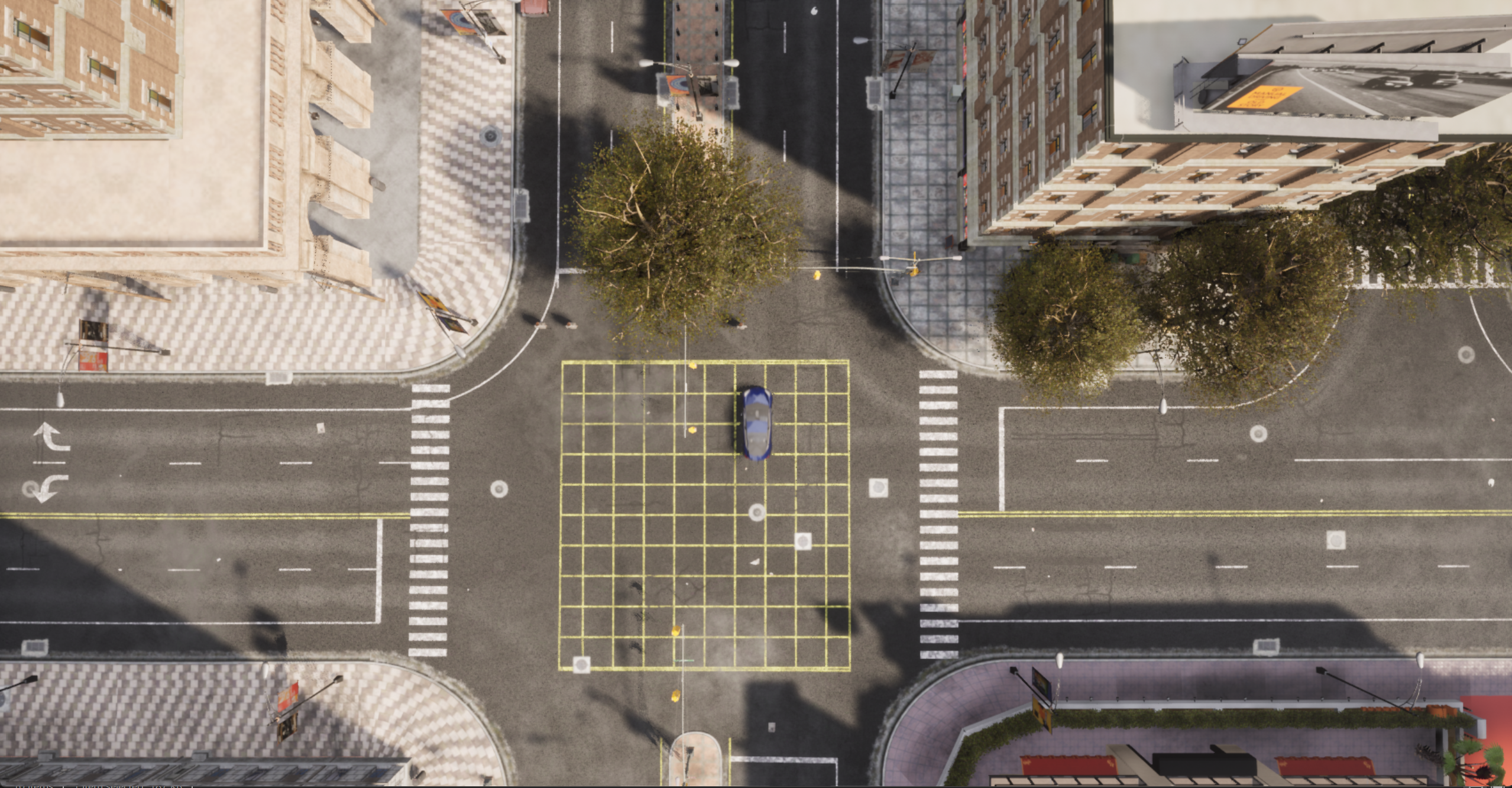}
    \caption{Abstract decision tree for the same scenario.}
  \end{subfigure}
  \caption{A representative collision-imminent scenario as rendered in CARLA
  for illustrative purposes (left) and its corresponding abstract decision
  tree (right), showing the three action branches and their pedestrian
  casualty counts under the Utilitarian framework. The CARLA render is
  provided for environmental context; the dataset itself consists of
  codebook-derived 40-dimensional state vectors aligned with the CARLA
  feature space.}
  \label{fig:scenario_pair}
\end{figure*}

The data is split 80/20 into 160 training and 40 test scenarios. Each
scenario is represented as a 40-dimensional state vector $s \in
\mathbb{R}^{40}$ encoding ego vehicle velocity, passenger count, lane
position, relative obstacle velocity, and pedestrian counts across all three
possible action directions, among other environmental features. These features
are aligned with those available within the CARLA simulation environment
\citep{dosovitskiy2017}, ensuring that the feature space remains
deployment-compatible with a full simulation pipeline in which a future
underlying AV model will be trained and scenarios will be generated. The CARLA
environment and a representative collision-imminent junction are illustrated
for context in \cref{fig:scenario_pair}.

\subsubsection{Label and Feedback Taxonomy}
\label{ssec:label_taxonomy}

Three distinct sources of human judgment inform this work and must be
distinguished clearly, as the central contribution depends on the contrast
between them.

\textbf{Codebook-derived normative labels} are determined deterministically
from the codebook rules for each scenario. The Utilitarian label selects the
action minimizing pedestrian casualties; the Kantian label uniformly selects
maintain course. These labels represent what each ethical framework
\emph{prescribes}, independent of any individual's judgment, and serve as
the ground-truth targets against which policy accuracy is measured.

\textbf{Human coder labels} are the action assignments produced by
\textit{2} independent graders applying the codebook rules to each scenario
during the iterative development rounds described below. These were used to
validate the codebook itself: disagreements between coders identified
ambiguities in the rule structure and drove revision of both the codebook and
the scenario set. They do not represent moral intuition; they represent the
reliability of rule application.

\textbf{Human preference feedback} is the pairwise annotation data used to
train the Bradley-Terry reward model. In a separate data collection, human
annotators were shown pairs of trajectories and asked which was more ethically
appropriate. This is the signal that the RLHF pipeline optimizes against. It
is categorically distinct from the codebook-derived labels. The central
finding of this paper is that this preference feedback diverged systematically
from what the codebook prescribed.

\begin{figure}[ht]
  \centering
  \includegraphics[width=0.8\columnwidth]{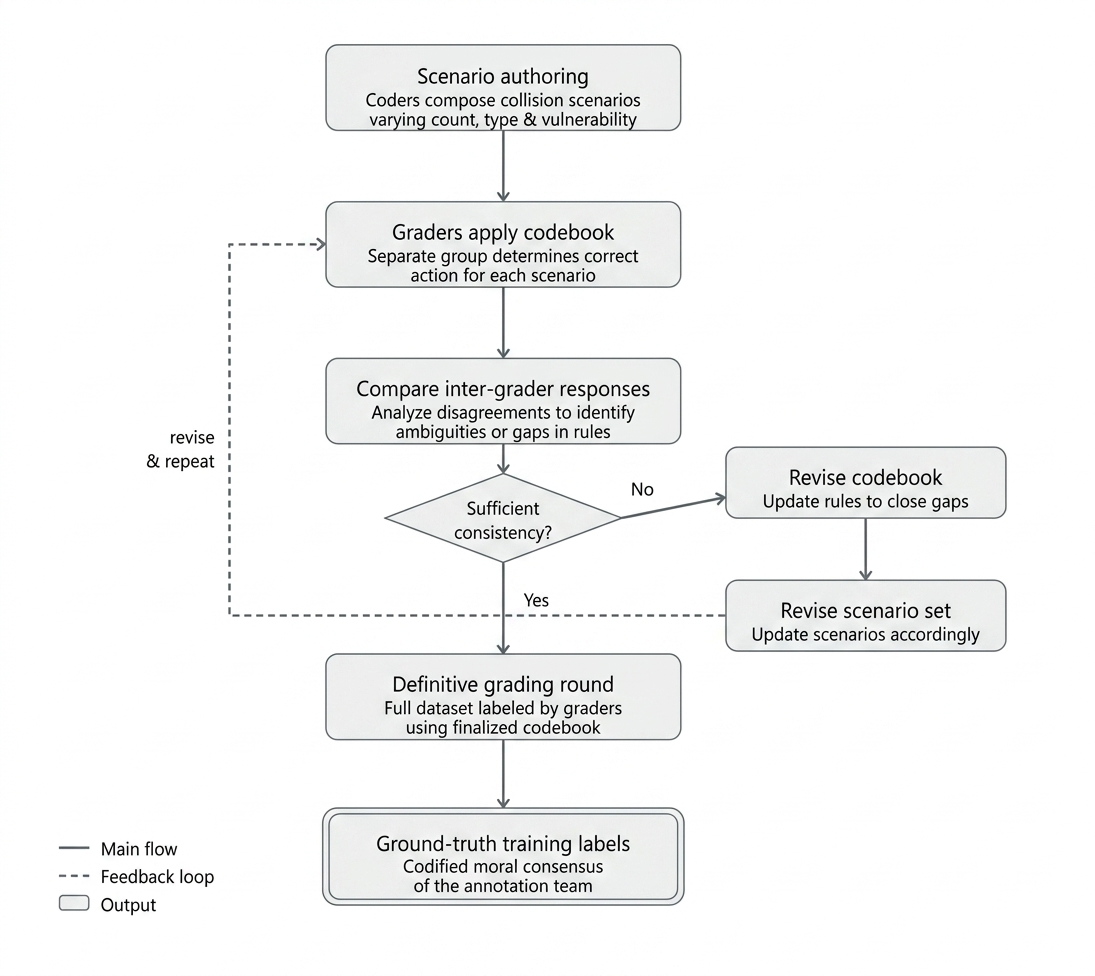}
  \caption{Iterative codebook development pipeline. Scenario authoring,
  grading, and inter-grader disagreement analysis cycle until sufficient
  consistency is reached, producing the ground-truth training labels.}
  \label{fig:codebook}
\end{figure}

\begin{figure}[ht]
  \centering
  \begin{subfigure}[t]{0.47\columnwidth}
    \includegraphics[width=\textwidth]{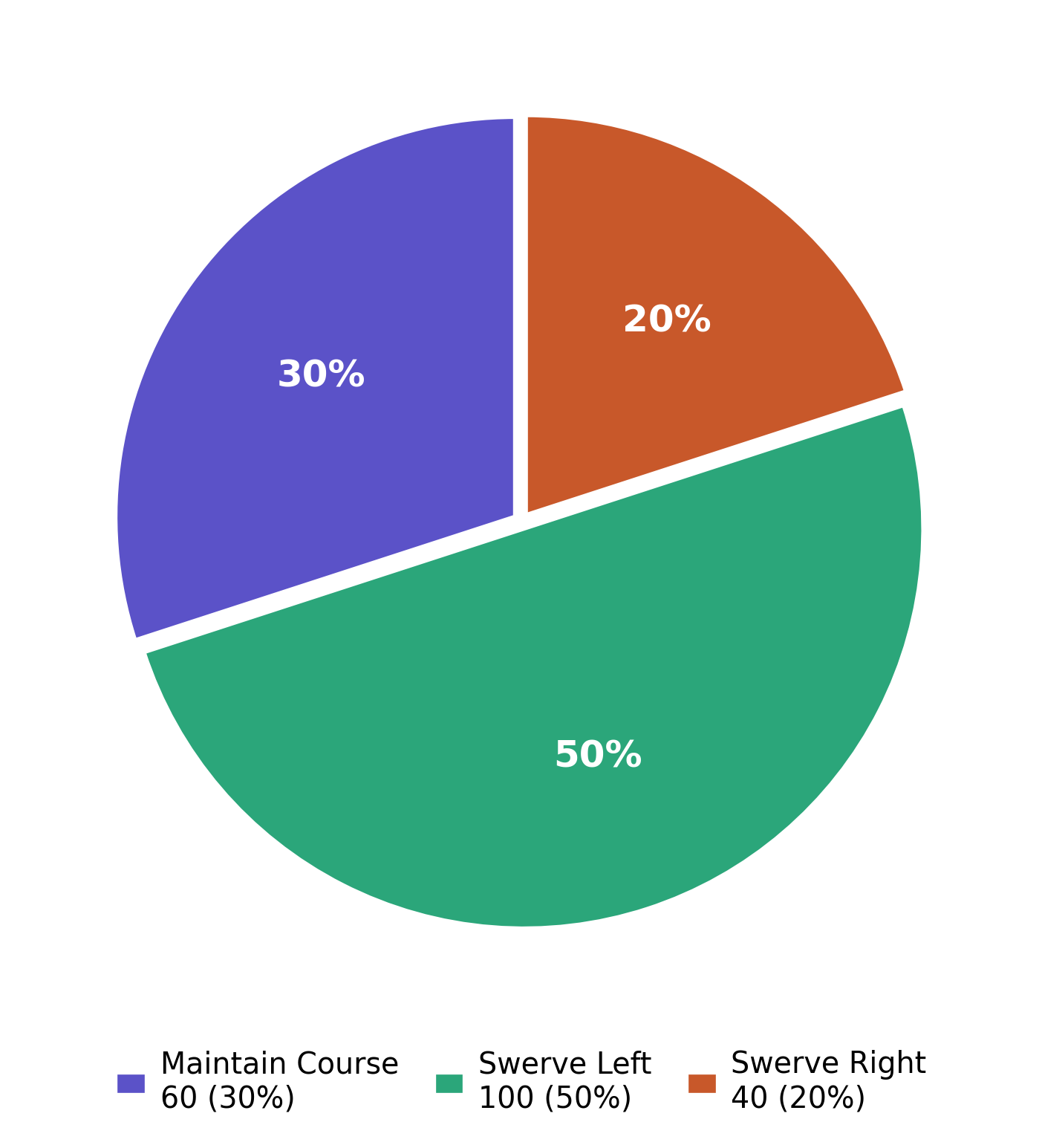}
    \caption{Label distribution (pie).}
  \end{subfigure}
  \hfill
  \begin{subfigure}[t]{0.44\columnwidth}
    \includegraphics[width=\textwidth]{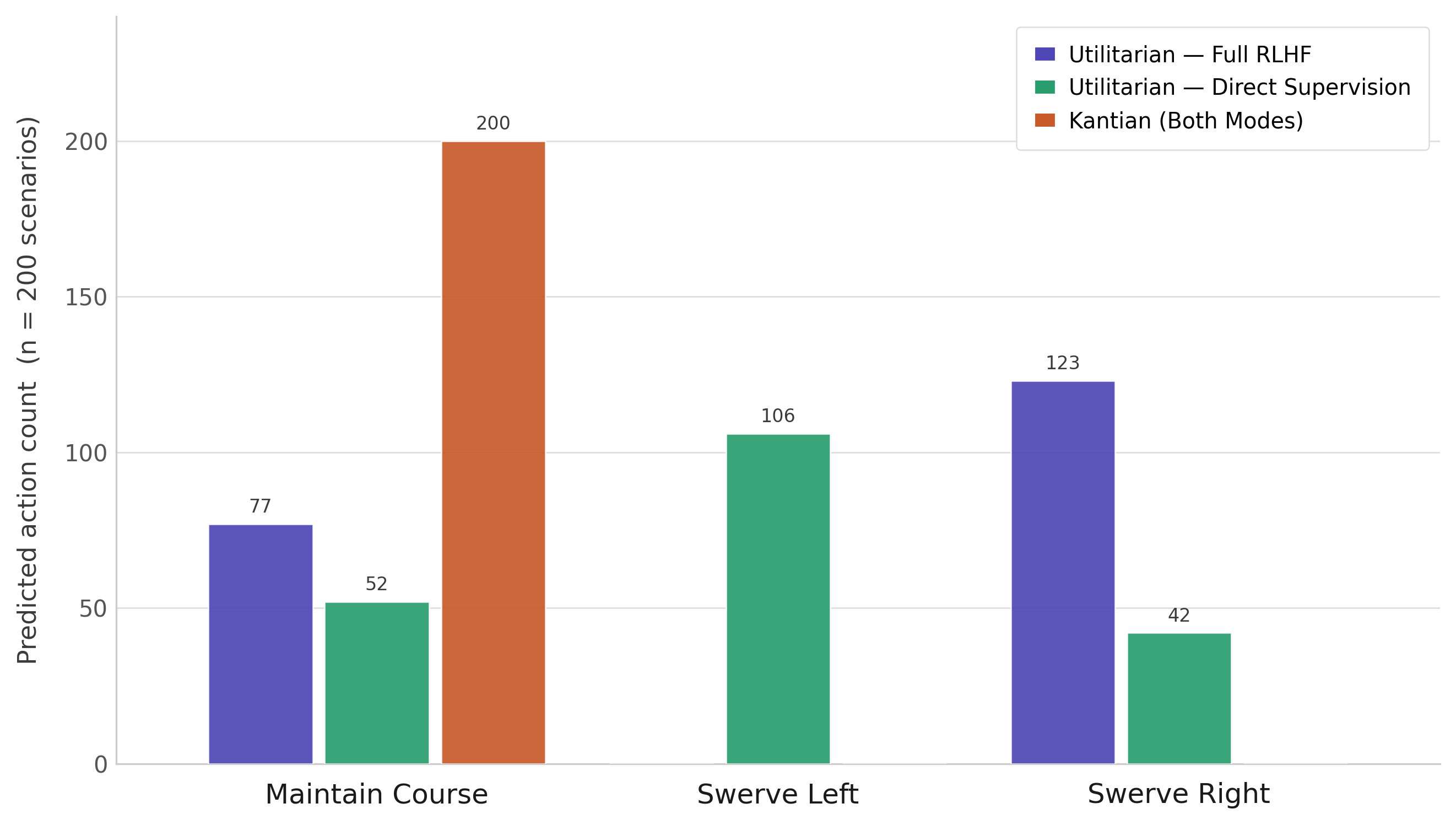}
    \caption{Label distribution (bar).}
  \end{subfigure}
  \caption{Ground-truth Utilitarian label distribution across all 200
  scenarios. Swerve-left constitutes 50\% of correct actions, making
  any policy that fails to learn this action incapable of exceeding
  50\% accuracy.}
  \label{fig:label_dist}
\end{figure}

Ground-truth labels are derived from the ethical framework under evaluation:
the Utilitarian label selects the action minimizing pedestrian casualties,
while the Kantian label uniformly selects maintain course across all
scenarios, reflecting the categorical imperative \citep{kant1785}. The
resulting Utilitarian label distribution is notably imbalanced: 60 maintain
course, 100 swerve-left, and 40 swerve-right, as illustrated in
\cref{fig:label_dist}, a property that proves consequential in later analysis.
Because swerve-left constitutes 50\% of correct Utilitarian actions, any
policy that fails to learn this action cannot exceed 50\% accuracy regardless
of its performance elsewhere; this ceiling makes the RLHF collapse reported in
the results particularly diagnostic.

The reliability of these ground-truth labels depends critically on the
codebook from which they are derived, whose iterative development process is
illustrated in \cref{fig:codebook}. The codebook was developed through a
structured iterative process in which human coders first composed a set of
scenarios describing unavoidable collision situations involving one or more
AVs, each varying the number, type, and vulnerability profile of individuals
across each possible action path. A separate group of human graders then
applied the codebook rules to determine the correct action for each scenario.
Following each grading round, inter-grader responses were compared and cases
of disagreement were analyzed to identify ambiguities or gaps in the rule
structure, with revisions applied to both the codebook and the scenario set
accordingly. This cycle was repeated until sufficient consistency was reached
across all scenarios \citep{krippendorff2004}, at which point a definitive
grading round was conducted over the full dataset. The resulting labeled
decisions constitute the ground-truth training labels for the EDH and
represent the codified moral consensus of the annotation team rather than the
judgment of any individual rater. Formal inter-rater reliability statistics
were not retained from the coding rounds; this is acknowledged as a limitation
in Section~\ref{sec:limitations}.

\subsubsection{Annotator Pool}
\label{ssec:annotators}

The human raters contributing preference feedback were two individuals
recruited from a general collegiate population. The sole inclusion criterion
was no prior formal exposure to normative ethical frameworks, specifically
Kantian deontology and Utilitarian theory, to ensure that preference
judgments reflected lay moral intuition rather than trained philosophical
reasoning. No additional demographic screening was applied. The human moral
intuition captured by the preference dataset therefore reflects the values of
a small, demographically specific group. Findings about what ``humans prefer''
should be interpreted as findings about what these two raters preferred;
generalizability to broader or more diverse populations is not claimed and is
acknowledged as a limitation in Section~\ref{sec:limitations}.

\subsection{Model Architecture}

\begin{figure}[ht]
  \centering
  \includegraphics[height=0.38\textheight]{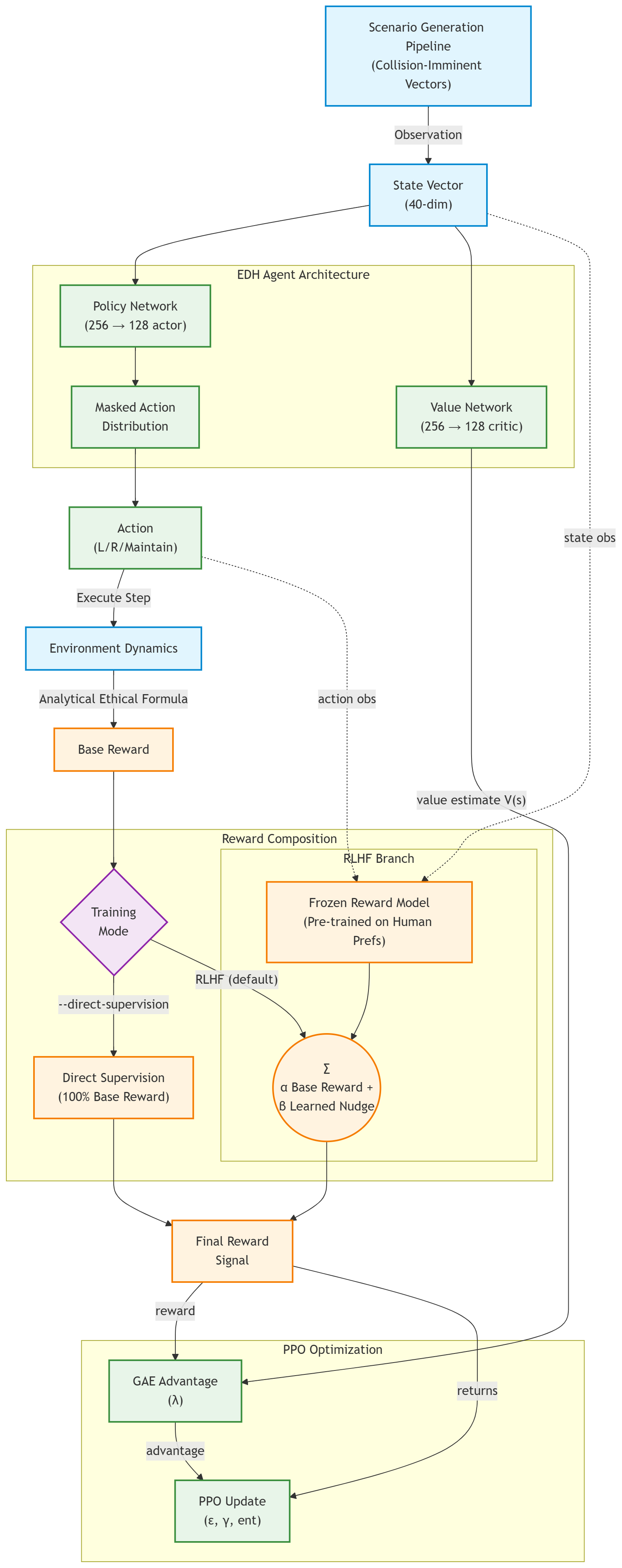}
  \caption{Full EDH pipeline architecture. The state vector flows through
  the policy and value networks, with reward composition branching between
  Full RLHF and Direct Supervision modes before PPO optimization.}
  \label{fig:architecture}
\end{figure}

With the scenario distribution and labeling protocol established, the
architecture that operates over this data can be defined, as illustrated in
\cref{fig:architecture}. The full pipeline integrates three learned components,
with each component serving a distinct role in translating raw state
observations into ethically-grounded action selections.

The \textbf{policy network} (actor) $\pi_\theta$ maps 40-dimensional states
to a probability distribution over five actions via a two-layer MLP
($256 \rightarrow 128$ hidden units), with actions brake and accelerate masked
out via large negative logit bias ($-10^9$), constraining the effective
decision space to $\mathcal{A} = \{\text{maintain, swerve-left,
swerve-right}\}$. The \textbf{value network} (critic) $V_\phi$ mirrors this
architecture, outputting a scalar state value used for Generalized Advantage
Estimation \citep{schulman2016gae}. The \textbf{reward model} $R_\psi$ takes
a concatenation of the state $s \in \mathbb{R}^{40}$ and a one-hot action
encoding $a \in \mathbb{R}^5$ as input (45 dimensions total) and outputs a
scalar ethical score per $(s, a)$ pair.

\subsubsection{Bradley-Terry Preference Model}
\label{ssec:rm}

The reward model is trained via the Bradley-Terry preference framework
\citep{bradleyterry1952} following the RLHF paradigm of
\citet{christiano2017}. Human annotators are shown pairs of trajectory
rollouts $(\tau_A, \tau_B)$ and indicate which trajectory executed the more
ethically appropriate sequence of actions. The cumulative reward assigned to a
trajectory $\tau$ is:
\[
  \hat{R}_\psi(\tau) \;=\; \sum_{(s,\, a)\,\in\, \tau} R_\psi(s, a)
\]
The probability that a human annotator prefers $\tau_A$ over $\tau_B$ is
modeled as:
\begin{equation}
  \label{eq:bt}
  P(\tau_A \succ \tau_B)
  \;=\; \sigma\!\left(\hat{R}_\psi(\tau_A) - \hat{R}_\psi(\tau_B)\right)
\end{equation}
The reward model parameters $\psi$ are updated by minimizing binary
cross-entropy over the preference dataset $\mathcal{D} = \{(\tau_A^{(i)},
\tau_B^{(i)}, y^{(i)})\}$, where $y^{(i)} = 1$ if annotators preferred
$\tau_A$:
\begin{multline}
  \label{eq:rm_loss}
  \mathcal{L}(\psi) \;=\;
  -\,\mathbb{E}_{(\tau_A,\tau_B,y)\sim\mathcal{D}}
  \bigl[
    y \log P(\tau_A \succ \tau_B) \\
    + (1 - y) \log\!\left(1 - P(\tau_A \succ \tau_B)\right)
  \bigr]
\end{multline}

\subsection{Training Configuration}

\subsubsection{Generalized Advantage Estimation}

Policy gradients are computed using Generalized Advantage Estimation (GAE)
\citep{schulman2016gae}, which reduces variance by combining multi-step
temporal difference estimates via exponential weighting. Given a rollout
buffer of $N = 2048$ transitions, the TD error at step $t$ is:
\begin{equation}
  \label{eq:td_error}
  \delta_t \;=\; r_t + \gamma\, V_\phi(s_{t+1})\,(1 - d_t) - V_\phi(s_t)
\end{equation}
where $d_t \in \{0, 1\}$ is a terminal-state indicator and $\gamma = 0.99$
is the discount factor. The advantage estimate is computed backward through
the buffer:
\begin{equation}
  \label{eq:gae}
  \hat{A}_t \;=\; \delta_t
    + (\gamma\lambda_{\text{GAE}})(1 - d_t)\,\hat{A}_{t+1}
\end{equation}
with GAE smoothing parameter $\lambda_{\text{GAE}} = 0.95$. Return targets
for critic regression are recovered as $G_t = \hat{A}_t + V_\phi(s_t)$.
Advantages are normalized to zero mean and unit variance prior to use in the
policy loss to stabilize gradient magnitudes across episodes of differing
reward scale.

\FloatBarrier
\subsubsection{PPO Clipped Surrogate Objective}

The actor policy $\pi_\theta$ is optimized using PPO \citep{schulman2017ppo}.
In the present benchmark each scenario constitutes a single-step decision,
reducing the formulation to a contextual bandit; PPO is retained for two
reasons. First, it provides architectural continuity with the intended
deployment target, a full end-to-end pipeline in which ethical decisions
unfold over multiple timesteps. Second, the clipped surrogate objective
explicitly bounds the magnitude of each policy update, preventing catastrophic
forgetting of behaviors encountered early in training, a property that proved
relevant under the high-variance reward signal of the Full RLHF condition.
Let the probability ratio between the updated and old policies be:
\[
  \rho_t(\theta) \;=\;
  \frac{\pi_\theta(a_t \mid s_t)}
       {\pi_{\theta_{\text{old}}}(a_t \mid s_t)}
\]
The actor loss combines the clipped objective with an entropy regularization
term, which prevents premature convergence to deterministic swerving
strategies:
\begin{multline}
  \label{eq:actor_loss}
  \mathcal{L}^{\text{Actor}}(\theta) \;=\;
  -\mathbb{E}_t\!\left[
    \min\!\Bigl(
      \rho_t \hat{A}_t,\;
      \operatorname{clip}(\rho_t,\,
        1{-}\varepsilon,\, 1{+}\varepsilon)\hat{A}_t
    \Bigr)
  \right] \\
  - c_{\text{ent}}\,\mathbb{E}_t\!\left[
    H\!\left(\pi_\theta(\cdot \mid s_t)\right)
  \right]
\end{multline}

with clipping threshold $\varepsilon = 0.2$ and entropy coefficient
$c_{\text{ent}} = 0.03$. The critic is updated by minimizing mean-squared
error against the return targets:
\begin{equation}
  \label{eq:critic_loss}
  \mathcal{L}^{\text{Critic}}(\phi)
  \;=\; \frac{1}{2}\,\mathbb{E}_t\!\left[
    \bigl(V_\phi(s_t) - G_t\bigr)^2
  \right]
\end{equation}
Gradients of the combined loss $\mathcal{L} = \mathcal{L}^{\text{Actor}} +
\mathcal{L}^{\text{Critic}}$ are clipped by global $\ell_2$ norm at $0.5$
before each Adam optimizer step to prevent destabilizing updates during
high-variance rollouts. The reward signal passed to the policy is the
composite $R_{\text{total}}$ defined in \cref{eq:reward_total}, where learned
rewards are z-score normalized and clipped to $[-3, 3]$ prior to mixing to
prevent value network instability. The reward model is updated every 50 PPO
iterations on a batch of human preference pairs, a frequency reduced from an
initial value of 10 to mitigate overfitting. \cref{tab:hyperparams}
summarizes all hyperparameters.

\begin{table}[ht]
\centering
\small
\caption{Training hyperparameters for the EDH.}
\label{tab:hyperparams}
\begin{tabular}{lll}
\toprule
\textbf{Parameter} & \textbf{Symbol} & \textbf{Value} \\
\midrule
Discount factor          & $\gamma$                  & $0.99$ \\
GAE smoothing            & $\lambda_{\text{GAE}}$    & $0.95$ \\
Rollout buffer size      & $N$                       & $2048$ \\
PPO epochs per rollout   & $K_{\text{ppo}}$          & $4$ \\
PPO clip threshold       & $\varepsilon$             & $0.2$ \\
Entropy coefficient      & $c_{\text{ent}}$          & $0.03$ \\
Policy LR                & $\eta_{\text{policy}}$    & $3 \times 10^{-4}$ \\
Gradient clip norm       &                           & $0.5$ \\
Reward model LR          & $\eta_{\text{rm}}$        & $10^{-3}$ \\
RM update frequency      &                           & $50$ \\
RM pref.\ batch size     & $B_{\text{pref}}$         & $32$ \\
Base reward weight       & $\alpha$                  & $0.3$ \\
Learned reward weight    & $\beta$                   & $0.7$ \\
Total timesteps          &                           & $2{,}000{,}000$ \\
\bottomrule
\end{tabular}
\end{table}

\FloatBarrier
\subsubsection{Training Configurations}

Two training configurations are evaluated. The \textbf{Full RLHF} mode trains
the policy and reward model jointly, with human preference pairs shaping the
gradient signal throughout. The \textbf{Direct Supervision} mode bypasses the
reward model entirely, training PPO directly on the shaped environment reward.
This ablation serves as a diagnostic, isolating whether observed failure modes
originate in the RLHF pipeline or the underlying environment formulation, a
distinction that proves critical in the results. At inference time, the policy
switches from stochastic sampling to a deterministic greedy strategy:
\[
  a_t \;=\; \arg\max_{a \in \mathcal{A}}\; \pi_\theta(a \mid s_t)
\]
This eliminates output variance while retaining the learned ethical weighting
acquired during training.

\FloatBarrier
\section{Results}

\subsection{Kantian Ethical Framework}

The Kantian condition serves as a methodological control rather than a
substantive ethical learning result. Under the codebook, the correct Kantian
action is uniformly ``maintain course'' across all 200 scenarios: a policy
that ignores the input entirely and always predicts ``maintain'' would achieve
the same outcome. The pipeline converged to this behavior reliably,
confirming that the training infrastructure is stable and that any failure to
learn a correct action cannot be attributed to optimization pathologies. This
result does not constitute evidence that the model learned Kantian ethical
reasoning; it confirms the pipeline functions as intended, and we report it as
such before proceeding to the substantive finding.

\subsection{Utilitarian Ethical Framework}

\begin{figure*}[!ht]
  \centering
  \includegraphics[width=0.70\textwidth]{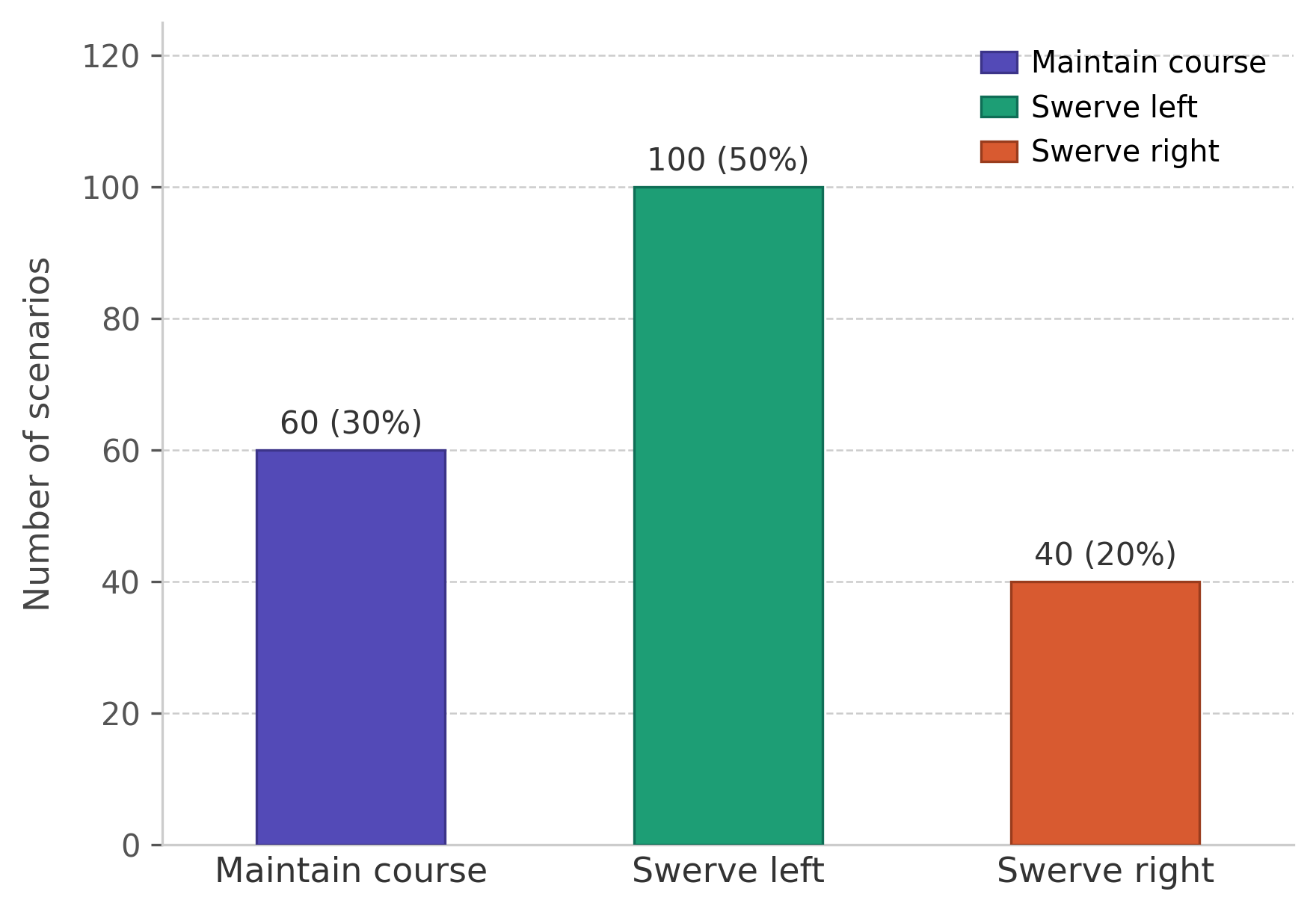}
  \caption{Predicted action distributions across all 200 scenarios under each
  training configuration. Under Full RLHF, the Utilitarian agent never
  predicts swerve-left, the modal correct action, collapsing instead toward
  maintain course and swerve-right. Direct Supervision recovers a balanced
  distribution aligned with ground-truth labels.}
  \label{fig:decision_dist}
\end{figure*}

The RLHF agent trained under the Utilitarian framework exhibited markedly
different behavior. At 2,000,000 time steps, the model achieved a testing
accuracy of only 42.5\%. An analysis of the decision distribution, illustrated
in \cref{fig:decision_dist}, revealed that in approximately 50\% of scenarios,
the agent deviated from strict casualty minimization in favor of
self-sacrifice, a behavior not prescribed by Utilitarian logic but one to
which human raters responded favorably. This divergence indicates that human
feedback introduced a systematic bias into the reward signal, pulling the
agent toward an emotionally salient but normatively inconsistent decision
pattern. The pattern is consistent with omission bias \citep{spranca1991,
ritov1990} and action aversion \citep{cushman2012}: human raters consistently
preferred inaction or self-sacrifice over active redirection of harm, even
when redirection would minimize total casualties.

\begin{figure*}[!ht]
  \centering
  \includegraphics[width=0.75\textwidth]{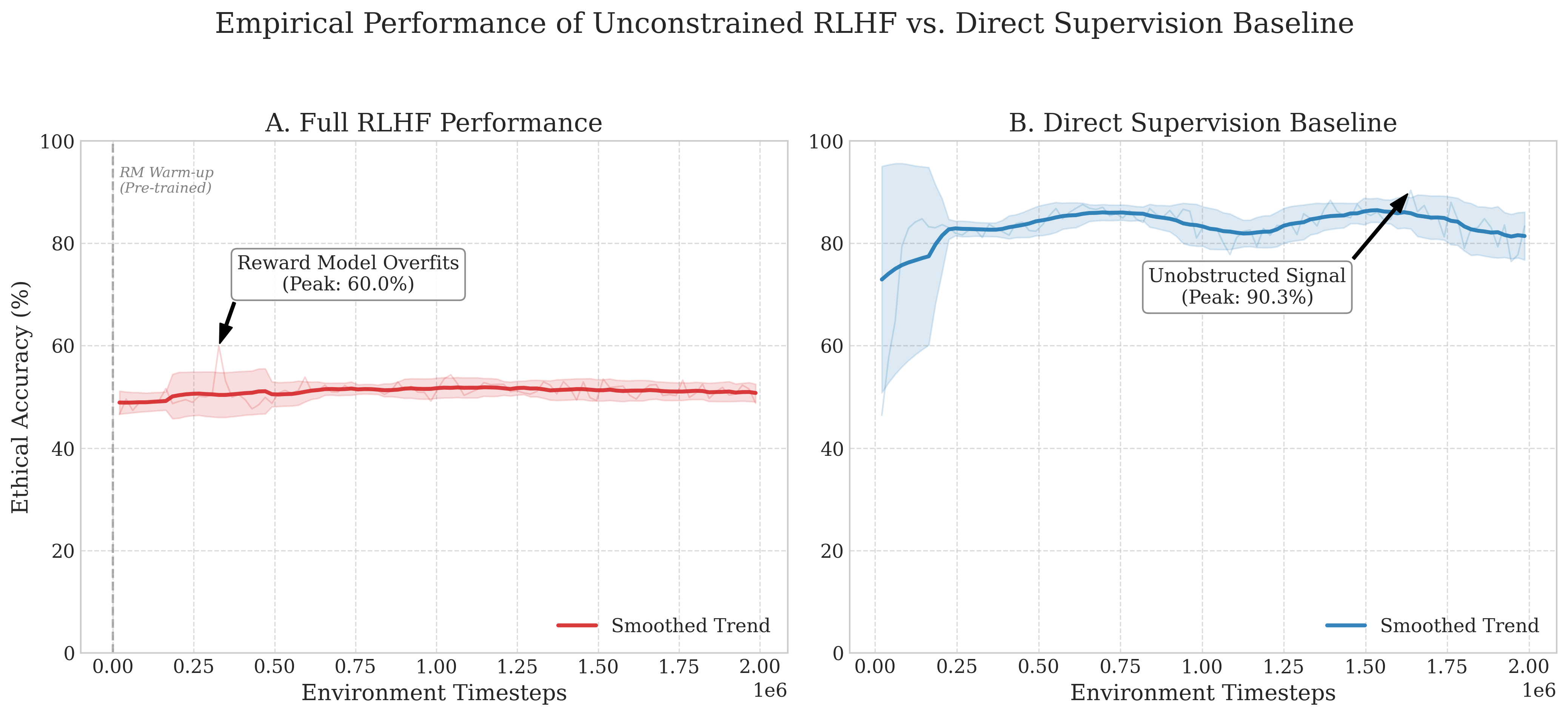}
  \caption{Training curves comparing Full RLHF (A) and Direct Supervision (B)
  under the Utilitarian framework over 2,000,000 environment timesteps. Panel
  A plateaus near the 50\% accuracy ceiling imposed by swerve-left collapse
  following reward model warm-up; Panel B reaches a peak of 90.3\%, confirming
  that the environment reward signal is well-formed and the failure mode is
  localized to the RLHF pipeline.}
  \label{fig:training_curves}
\end{figure*}

\begin{figure}[ht]
  \centering
  \includegraphics[width=\columnwidth]{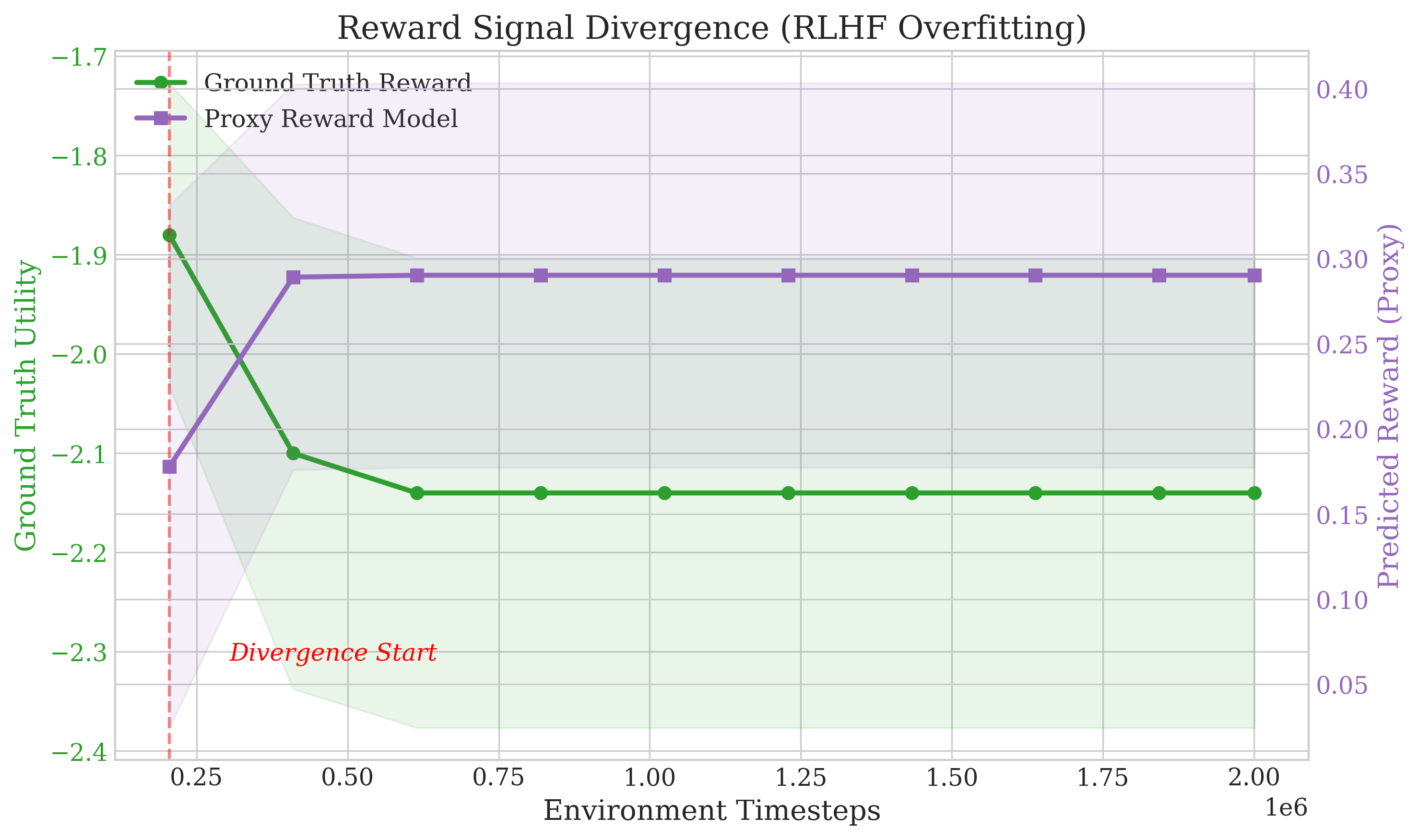}
  \caption{Reward signal divergence under Full RLHF. Following the divergence
  point the proxy reward model saturates while ground truth utility continues
  to decline, confirming reward model overfitting to human preference noise
  rather than the normative objective.}
  \label{fig:reward_divergence}
\end{figure}

The training dynamics underlying this collapse are shown in
\cref{fig:training_curves}, and the divergence between the proxy and
ground-truth reward signals is plotted in \cref{fig:reward_divergence}. To
isolate the effect of this bias, human feedback was removed from the training
loop, reducing the architecture to the Direct Supervision configuration. This
modification produced a substantial improvement, with the model reaching a
peak accuracy of 90.3\%. These results confirm that
Utilitarianism is learnable under direct supervision, but that its formal
structure is sensitive to perturbation from human preferences that do not
align with strict casualty minimization.

\section{Discussion and Limitations}

\subsection{Discussion}

The results reveal a fundamental asymmetry in the learnability of classical
ethical frameworks when human preferences are introduced into the training
loop. The Kantian condition converged reliably to the single prescribed action,
functioning as a pipeline control. The substantive finding lies entirely with
the Utilitarian agent.

The Utilitarian framework exposed an inherent tension between formal normative
theory and empirical human moral intuition. The emergent self-sacrificial
behavior observed in the RLHF condition is not an artifact of model failure;
it is a reflection of something more consequential. Human raters, in
attempting to reward moral behavior, responded to perceived heroism rather
than strict outcome evaluation, effectively teaching the agent a coherent but
normatively incorrect objective \citep{greene2001}. This pattern is consistent
with omission bias \citep{spranca1991, ritov1990} and action aversion
\citep{cushman2012}: a robust finding in moral psychology that agents judge
harmful outcomes more harshly when they result from active intervention than
from inaction, even when the intervention would reduce total harm. Our
contribution is to demonstrate that this bias does not merely describe how
humans reason about moral scenarios in the abstract; it propagates faithfully
through an RLHF pipeline, shaping the learned policy of a safety-critical
decision system.

This divergence between what humans prescribe in theory and what they reward
in practice \citep{bonnefon2016} is precisely the dynamic this paper set out
to investigate. In this sense, the model did not fail to learn ethics; it
learned human ethics as humans practice it, not as philosophers define it.

The recovery to high accuracy upon removal of human feedback lends empirical
support to this interpretation and raises a broader design question: when the
goal is the faithful instantiation of a specific normative theory, is human
supervision an appropriate alignment mechanism at all? This is a
counterintuitive result, given that RLHF was originally motivated by the
desire to align model behavior with human values \citep{christiano2017}, yet
the findings here suggest that human values and normative ethical theory are
not always the same thing. Conflating the two in system design carries
measurable consequences.

\subsection{Limitations}
\label{sec:limitations}

Several limitations are worth noting. First, the omission bias observed under
the Utilitarian condition is likely one instance of a broader class of
affective biases that human raters bring to morally charged scenarios,
including loss aversion \citep{kahneman1979} and in-group favoritism
\citep{cushman2012}. Future work should investigate whether rater calibration,
structured annotation protocols, or hybrid feedback mechanisms can mitigate
these distortions without abandoning the human feedback paradigm entirely.

Second, the ethical frameworks evaluated here, Kantianism and Utilitarianism,
do not exhaust the space of normative theories relevant to AV
decision-making. Frameworks such as virtue ethics and contractualism, which
integrate both rule-based and consequentialist reasoning, remain unaddressed
\citep{gogoll2017, awad2018}. Whether the learnability asymmetry reported here
generalizes across such frameworks is an open empirical question.

Third, formal inter-rater reliability statistics were not retained from the
iterative codebook development process. The codebook was refined until
inter-grader consistency was reached \citep{krippendorff2004}, but no
Krippendorff's alpha or Cohen's kappa value was recorded. We acknowledge this
as a methodological limitation; future iterations of this pipeline should
instrument the coding rounds to capture and report these statistics. If the
raw grading data is recoverable, computing and reporting these values should
be treated as a priority revision.

Fourth, the preference feedback was collected from two raters drawn from a
collegiate population with no prior exposure to normative ethics. The
propagation of omission bias through the RLHF pipeline is therefore a finding
about this specific pair's preferences. Whether the same pattern holds across
larger or more diverse rater pools is an open empirical question, and the
small sample size means individual variation cannot be separated from
population-level effects. Two annotators also represents a minimal sample for
a Bradley-Terry preference model, which relies on aggregated pairwise
comparisons to estimate a stable reward signal; with two raters, the learned
reward model reflects a narrow preference distribution and may be sensitive to
idiosyncratic judgments that a larger pool would average out. Future work
should replicate the preference collection at scale before drawing strong
conclusions about the generalizability of the observed RLHF collapse.

Finally, the evaluation dataset of 200 moral edge cases, while diverse in its
coverage of the codebook-defined variable space, represents a finite
approximation of the moral distribution an AV would encounter in deployment.
The degree to which this benchmark predicts ethical behavior in practice
remains an open question, one that large-scale simulation or real-world data
collection would be required to answer.


\FloatBarrier
\FloatBarrier

\bibliographystyle{plainnat}

\begin{thebibliography}{99}

\bibitem[Bentham(1789)]{bentham1789}
Bentham, J. (1789).
\newblock \textit{An Introduction to the Principles of Morals and Legislation}.
\newblock T. Payne and Son, London.

\bibitem[Bonnefon et al.(2016)]{bonnefon2016}
Bonnefon, J.-F., Shariff, A., and Rahwan, I. (2016).
\newblock The social dilemma of autonomous vehicles.
\newblock \textit{Science}, 352(6293):1573--1576.

\bibitem[Bradley and Terry(1952)]{bradleyterry1952}
Bradley, R.~A. and Terry, M.~E. (1952).
\newblock Rank analysis of incomplete block designs: I. The method of paired
comparisons.
\newblock \textit{Biometrika}, 39(3/4):324--345.

\bibitem[Christiano et al.(2017)]{christiano2017}
Christiano, P., Leike, J., Brown, T.~B., Martic, M., Legg, S., and Amodei,
D. (2017).
\newblock Deep reinforcement learning from human preferences.
\newblock In \textit{Advances in Neural Information Processing Systems
(NeurIPS)}.

\bibitem[Cushman et al.(2012)]{cushman2012}
Cushman, F., Gray, K., Gaffey, A., and Mendes, W.~B. (2012).
\newblock Simulating murder: The aversion to harmful action.
\newblock \textit{Emotion}, 12(1):2--7.

\bibitem[Dosovitskiy et al.(2017)]{dosovitskiy2017}
Dosovitskiy, A., Ros, G., Codevilla, F., Lopez, A., and Koltun, V. (2017).
\newblock CARLA: An open urban driving simulator.
\newblock In \textit{Proceedings of the 1st Annual Conference on Robot
Learning (CoRL)}.

\bibitem[Foot(1967)]{foot1967}
Foot, P. (1967).
\newblock The problem of abortion and the doctrine of double effect.
\newblock \textit{Oxford Review}, 5:5--15.

\bibitem[Gabriel(2020)]{gabriel2020}
Gabriel, I. (2020).
\newblock Artificial intelligence, values, and alignment.
\newblock \textit{Minds and Machines}, 30(3):411--437.

\bibitem[Greene et al.(2001)]{greene2001}
Greene, J.~D., Sommerville, R.~B., Nystrom, L.~E., Darley, J.~M., and Cohen,
J.~D. (2001).
\newblock An fMRI investigation of emotional engagement in moral judgment.
\newblock \textit{Science}, 293(5537):2105--2108.

\bibitem[Kahneman and Tversky(1979)]{kahneman1979}
Kahneman, D. and Tversky, A. (1979).
\newblock Prospect theory: An analysis of decision under risk.
\newblock \textit{Econometrica}, 47(2):263--291.

\bibitem[Kant(1785)]{kant1785}
Kant, I. (1785).
\newblock \textit{Groundwork of the Metaphysics of Morals}.
\newblock Translated by M. Gregor. Cambridge University Press, Cambridge.

\bibitem[Mill(1863)]{mill1863}
Mill, J.~S. (1863).
\newblock \textit{Utilitarianism}.
\newblock Parker, Son, and Bourn, London.

\bibitem[Ritov and Baron(1990)]{ritov1990}
Ritov, I. and Baron, J. (1990).
\newblock Reluctance to vaccinate: Omission bias and ambiguity.
\newblock \textit{Journal of Behavioral Decision Making}, 3(4):263--277.

\bibitem[Russell(2019)]{russell2019}
Russell, S. (2019).
\newblock \textit{Human Compatible: Artificial Intelligence and the Problem of
Control}.
\newblock Viking, New York.

\bibitem[SAE International(2018)]{sae2018}
SAE International (2018).
\newblock Taxonomy and definitions for terms related to driving automation
systems for on-road motor vehicles.
\newblock \textit{SAE Standard J3016}.

\bibitem[Schulman et al.(2016)]{schulman2016gae}
Schulman, J., Moritz, P., Levine, S., Jordan, M., and Abbeel, P. (2016).
\newblock High-dimensional continuous control using generalized advantage
estimation.
\newblock In \textit{International Conference on Learning Representations
(ICLR)}.

\bibitem[Schulman et al.(2017)]{schulman2017ppo}
Schulman, J., Wolski, F., Dhariwal, P., Radford, A., and Klimov, O. (2017).
\newblock Proximal policy optimization algorithms.
\newblock \textit{arXiv preprint arXiv:1707.06347}.

\bibitem[Spranca et al.(1991)]{spranca1991}
Spranca, M., Minsk, E., and Baron, J. (1991).
\newblock Omission and commission in judgment and choice.
\newblock \textit{Journal of Experimental Social Psychology}, 27(1):76--105.

\bibitem[Stiennon et al.(2020)]{stiennon2020}
Stiennon, N., Ouyang, L., Wu, J., Ziegler, D.~M., Lowe, R., Voss, C.,
Radford, A., Amodei, D., and Christiano, P. (2020).
\newblock Learning to summarize with human feedback.
\newblock In \textit{Advances in Neural Information Processing Systems
(NeurIPS)}.

\bibitem[Paden et al.(2016)]{paden2016}
Paden, B., Cap, M., Yong, S.~Z., Yershov, D., and Frazzoli, E. (2016).
\newblock A survey of motion planning and control techniques for self-driving
urban vehicles.
\newblock \textit{IEEE Transactions on Intelligent Vehicles}, 1(1):33--55.

\bibitem[Krippendorff(2004)]{krippendorff2004}
Krippendorff, K. (2004).
\newblock \textit{Content Analysis: An Introduction to Its Methodology}.
\newblock Sage, Thousand Oaks, CA.

\bibitem[Awad et al.(2018)]{awad2018}
Awad, E., Dsouza, S., Kim, R., Schulz, J., Henrich, J., Shariff, A.,
Bonnefon, J.-F., and Rahwan, I. (2018).
\newblock The Moral Machine experiment.
\newblock \textit{Nature}, 563(7729):59--64.

\bibitem[Gogoll and M\"{u}ller(2017)]{gogoll2017}
Gogoll, J. and M\"{u}ller, J.~F. (2017).
\newblock Autonomous cars: In favor of a mandatory ethics setting.
\newblock \textit{Science and Engineering Ethics}, 23(3):681--700.

\end{thebibliography}

\end{document}